\documentclass{article}
\usepackage{spconf,amsmath,graphicx}
\usepackage{todonotes}
\usepackage{color}
\usepackage{subcaption}
\graphicspath{{fig/}}

\title{ {Magnetic Resonance Fingerprinting using Recurrent Neural Networks}}

\name{Ilkay Oksuz$^{1}$\thanks{This work was supported by an EPSRC programme Grant (EP/P001009/1) and the Wellcome EPSRC Centre for Medical Engineering at School of Biomedical Engineering and Imaging Sciences, King’s College London (WT 203148/Z/16/Z). We acknowledge financial support from the Department of Health via the NIHR comprehensive Biomedical Research Centre award to Guys \& St Thomas NHS Foundation Trust with KCL and Kings College Hospital NHS Foundation Trust.},  
Gastao Cruz$^{1}$,  
James Clough$^{1}$,  
Aurelien Bustin$^{1}$, 
Nicolo Fuin$^{1}$, 
Rene M. Botnar$^{1}$,  
}
\secondlinename{Claudia Prieto$^{1*}$, Andrew P. King$^{1*}$,  
Julia A. Schnabel$^{1*}$ \thanks{$^{*}$ Joint last authors.}}

\address{$^{1}$School of Biomedical Engineering \& Imaging Sciences , King\rq{}s College London, UK 
}

\begin{document}

\maketitle
\begin{abstract}

Magnetic Resonance Fingerprinting (MRF) is a new approach to quantitative magnetic resonance imaging that allows simultaneous measurement of multiple tissue properties in a single, time-efficient acquisition. Standard MRF reconstructs parametric maps using dictionary matching and lacks scalability due to computational inefficiency. We propose to perform MRF map reconstruction using a recurrent neural network, which exploits the time-dependent information of the MRF signal evolution. We evaluate our method on multiparametric synthetic signals and compare it to existing MRF map reconstruction approaches, including those based on neural networks. Our method achieves state-of-the-art estimates of T1 and T2 values. In addition, the reconstruction time is significantly reduced compared to dictionary-matching based approaches.

\end{abstract}
\begin{keywords}
Magnetic resonance fingerprinting, Recurrent Neural Networks, LSTM, GRU, Parameter mapping
\end{keywords}

\section{Introduction}
\label{sec:intro}

Magnetic resonance imaging (MRI) is a powerful and versatile non-invasive imaging modality, which is widely used for the diagnosis of pathologies. However, most MRI techniques are restricted to qualitative imaging and are not capable of characterizing tissue in an objective and reproducible fashion.  To overcome this limitation, magnetic resonance fingerprinting (MRF) has recently been proposed \cite{Ma2013}.

\begin{figure*}[htb]
 \centering
  \centerline{\includegraphics[width=16.0cm]{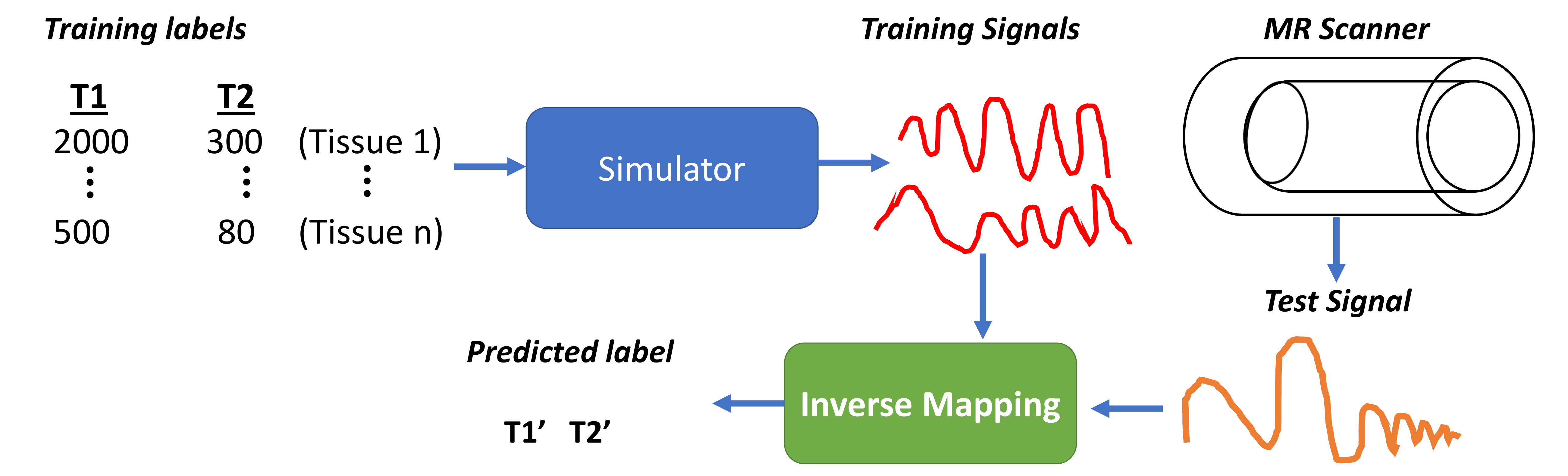}}
%
\caption{Dictionary-matching based MRF map reconstruction methods. The goal is to match the measured fingerprint with the pre-calculated dictionary signal using an inverse mapping. We propose to use a recurrent neural network architecture for training our algorithm and predict the T1 and T2 values with the trained network.}
\label{fig:Motivation}
\end{figure*}

MRF scans an object multiple times with varying experimental parameters, and generates unique signal evolutions (or \emph{fingerprints}) as a function of the multiple material properties under investigation.
The most likely material of each image pixel is estimated by matching its measured fingerprint with a set of pre-calculated fingerprints in a dictionary, and choosing the one with the maximum dot-product. 
The MRF acquisition is designed as a simultaneous function of multiple tissue parameters and therefore several MR parameters can be reconstructed from a single acquisition. Quantitative maps of these parameters are produced from the results of the dictionary matching in a pixel-wise fashion \cite{Coppo2016} (see Fig. \ref{fig:Motivation}). However, this dictionary matching process is time-consuming, lacks scalability, and can introduce artefacts due to under-sampling of k-space during the MRF data acquisition \cite{Wang2014}.

\section{Related Works}
\label{sec:related}

 Early works focused on dictionary matching to improve the efficiency of inner product calculation and introduce spatial constraints. 
 Gomez et al. \cite{Gomez2015} proposed a spatial dictionary matching technique that matches a spatial neighborhood of fingerprints instead of using a  pixel-wise approach. 
 However, using a search window comes at the cost of requiring spatially aligned MRF time-point images, and lacks scalability, which is a general issue with dictionary-matching based MRF map reconstruction methods.
 Recently, several techniques have been proposed to overcome the limitations of dictionary matching in MRF reconstruction, particularly neural networks with the capability to address the scalability bottleneck.
 Fully-connected neural networks \cite{Cohen2018}, convolutional neural networks (CNNs) \cite{Hoppe2017} and complex valued networks \cite{Virtue2017} have all been proposed to learn the matching of MRF signal evolutions to nuclear magnetic resonance properties. 
 In addition, Balsiger et al. \cite{Balsiger2018} proposed the use of spatially aware 2D convolutions on brain MR scans to incorporate spatial information into the reconstructed MRF parameter map; and Fang et al. \cite{Fang2018} proposed a U-net architecture with relative difference based loss function to generate MRF parameter maps. 
 All of these approaches show promising results in terms of reconstruction accuracy and speed.
 However, none of them take the temporal nature of the signal evolution into account. 
 The variation of MR parameter values in the MRF signal generation is continuous, so there is some temporal redundancy in the MRF data. 
 More importantly, the pattern actually repeats through time and consequently the magnetization is led through similar contrasts.

In this paper, we propose a MRF reconstruction approach that exploits the temporal redundancy of signal evolutions. 
Our approach is based on recurrent neural networks (RNNs) and yields faster and more accurate estimates of T1 and T2 values than recently proposed methods on synthetic datasets. 
We compare our performance to dictionary matching as well as existing deep learning-based methods. 
We report quantitative results and discuss the potential of RNN-based approaches for accurate and fast MRF reconstruction.

\section{Methods}
\label{sec:format}

In this section, we provide an overview of our recurrent neural network architecture that we use for MRF label prediction. We give an overview of RNNs. In particular, we describe Long Short Term Memory (LSTM) and Gated Recurrent Units (GRU) with an emphasis on their differences with traditional RNNs.

\subsection{RNN Model for T1 and T2 label prediction}
\begin{figure}[htb]
 \centering
  \centerline{\includegraphics[width=9.0cm]{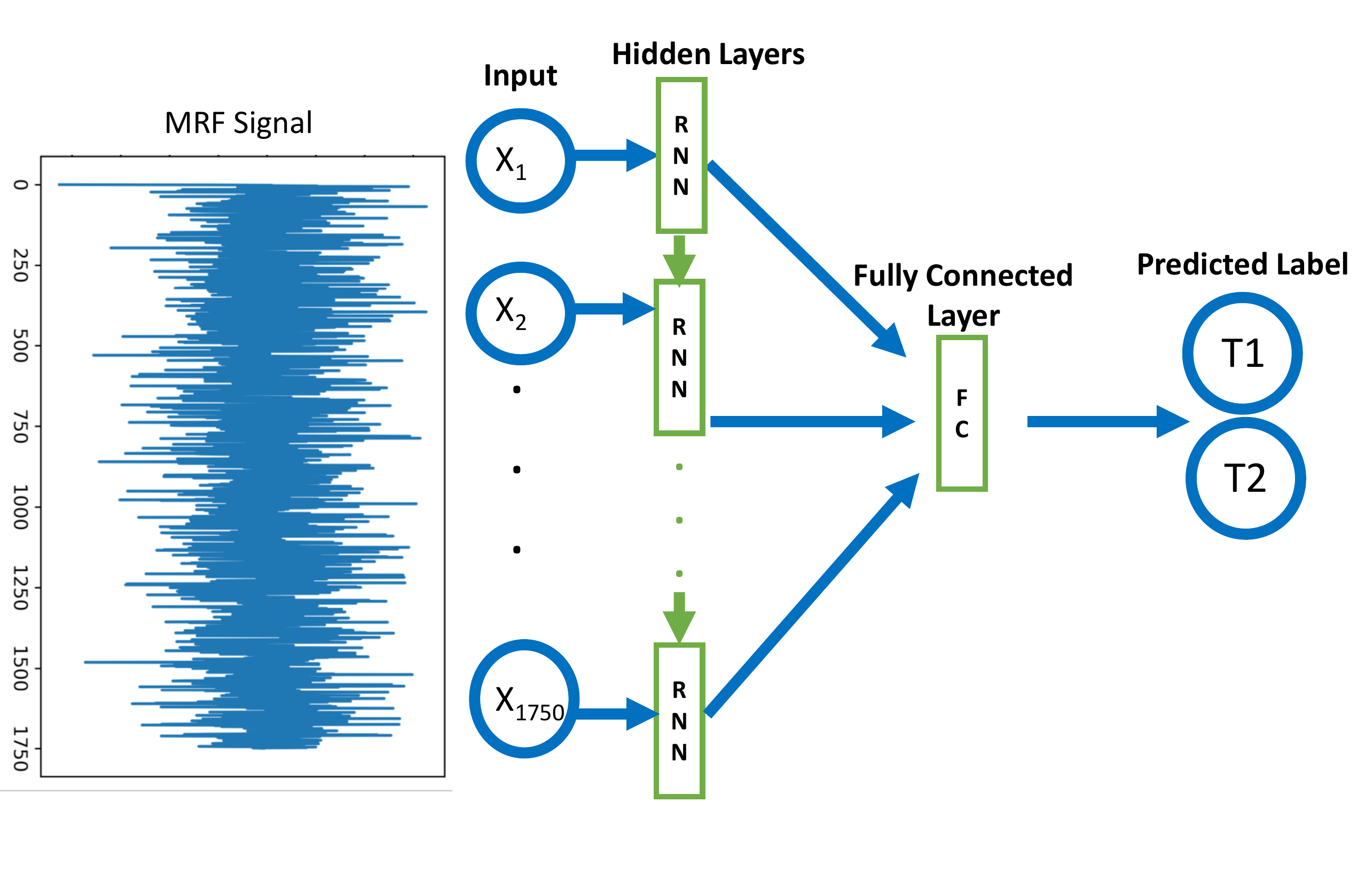}}
%
\caption{MR Fingerprinting dictionary matching signals using Recurrent Neural Networks. The input signal has 1750 values. 100 recurrent blocks are used for training and connected to a fully connected layer with 2 nodes for prediction of T1 and T2 values.}
\label{fig:RNN}
\end{figure}

We propose a method that uses RNNs for predicting the T1 and T2 labels using a dictionary of training data as visualized in Fig. \ref{fig:RNN}. 
Our network consists of recurrent blocks and a fully connected layer with 2 nodes for prediction of T1 and T2 values.

In RNN models, the input data are fed to the network one by one and the nodes in the network store their state at one time step and use it to inform the next time step. Unlike classical neural networks, RNNs use temporal information from the input data, which make them more appropriate for time series data. 
A RNN realizes this ability by recurrent connections between the neurons. A general equation for the RNN hidden state $h_t$, given an input sequence $ x=(x_{1}, x_{2},\ldots,x_{T} )$ is:

\begin{equation}
    h_t= \begin{cases}
    0,& \text{if } (t=0)\\  
    \phi (h_{t-1},x_{t}),         & \text{otherwise}
\end{cases}
\end{equation}
where $\phi$ is a non-linear function. The update of the recurrent hidden state is defined by:

\begin{equation}
h_{t}= g (Wx_{t}+ U h_{t-1})
\end{equation}
where g is a hyperbolic tangent function.

In general, this formulation of RNNs without memory cells suffers from vanishing gradient problems.
To address this issue, in this paper, we investigate the performance of two types of RNNs that incorporate memory cells, namely LSTM and GRU.

\subsection{Long Short Term Memory}
\label{sec:GRU}

The LSTM \cite{Hochreiter1997} layer is a recurrent layer designed to learn to store information over long time scales.
LSTMs have their own cell state. Whereas normal RNNs take in their previous hidden state and the current input, and output a new hidden state, an LSTM also takes in its old cell state and outputs its new cell state. This property helps LSTMs to address the vanishing gradient problem.

 An LSTM has three gates: an input gate, a forget gate  and an output gate. A sigmoid function is applied to the inputs  and the previous hidden state $h_{t-1}$. The goal of the LSTM is to generate the current hidden state at time $t$.

\subsection{Gated Recurrent Units}
\label{sec:GRU}

A GRU \cite{Cho2014} has two gates: a reset gate and an update gate. The update gate defines how much of the previous memory is to be kept and the reset gate determines how to combine the new input with the previous memory. GRUs become equivalent to standard RNNs if the reset gates are all 1 and the update gates all 0. The activation of the GRU at time t is a linear interpolation between the previous activation $h_{t-1}$ and the candidate activation $h_{t}$.

This procedure of taking a linear sum between the existing state and the newly computed state is similar to the LSTM unit. Unlike LSTM, GRU does not have any control over the state that is exposed, but exposes the whole state each time.

GRUs have the same fundamental idea of a gating mechanism to learn long-term dependencies as LSTM, but are different in several respects. 
First, a GRU has two gates and fewer parameters compared to LSTM. 
Second, GRUs do not possess any internal memory that is different from the exposed hidden state. Third, LSTMs have output gates and GRUs do not possess output gates. 
Finally, in LSTMs there is a second non-linearity applied when computing the output, which is not present in GRUs.

\subsection{Implementation details}

Our network has a visible layer with 1750 inputs (the length of the MRF signal), a hidden layer with 100 recurrent blocks or neurons, and a fully connected output layer that makes T1 and T2 value predictions. The default sigmoid activation function is used for the recurrent blocks. 

In this study, we use the Adam optimizer to minimize the mean square error loss. During training, a batch-size of 50 signals was used. The learning rate was 0.0001. The training ends when the network does not significantly improve its performance on the validation set for a predefined number of epochs (100). An improvement is considered significant if the relative increase in performance is at least 0.5\%.

\begin{table*}[htb]
\centering
\caption{Mean absolute error (MAE) and root mean square error (RMSE) for the T1 and T2 map values in ms. Mean and standard deviation values are reported for each method. Computational time of single signal prediction is reported for each method in ms.}
\label{tab:ED_table}
\begin{tabular}{lccccc}
\hline
 & \multicolumn{2}{c}{T1} & \multicolumn{2}{c}{T2} \\ 
\cline{2-3}
\cline{4-5}
Methods   & MAE & RMSE & MAE & RMSE & Computational Time \\
\hline 

ANN  \cite{Cohen2018}       & $113.22\pm39.59 $  & $ 119.94\pm46.48   $ & $ 67.16\pm54.71   $    & $86.63\pm103.36$ & 18 ms \\
1D CNN  \cite{Hoppe2017}    &$108.98\pm27.42 $ & $ 112.38\pm47.79  $ & $ 64.57\pm53.09   $ & $ 113.22\pm39.59$ & 14ms  \\
Inner Product               & $212.13\pm25.23 $  & $ 213.83\pm103.36 $ & $ 163.14\pm75.97 $  & $ 83.60\pm 49.03$ & 127 ms \\ 
\hline
\textbf{LSTM}               &$107.89\pm17.87 $  & $ 109.36\pm41.99  $ & $ 62.20\pm22.05   $  & $ 81.11\pm46.32$ & 15 ms  \\
\textbf{GRU}                &$101.81\pm17.17 $ & $ 103.25\pm39.15  $ & $ 58.91\pm14.21   $  & $ 73.66\pm34.39 $ & 14 ms \\
\hline

\end{tabular}
\label{tab:Quan}
\end{table*}

\section{Experimental Results}
\label{sec:Results}

We evaluated our algorithm on a  synthetically generated dataset of complex MRF signals. The MRF dictionary was generated using the Extended Phase Graph (EPG) formalism \cite{Hennig2004} for a range of T1 = [0:2:500] [500:5:1000] [1000:10:2000] [2000:50:4000]ms, T2 = [0:1:100] [100:2:500] ms. A gradient echo readout is utilised with a fixed repetition time (TR) = 4.3 ms. Note that dictionary entries where T2$<$T1 were not simulated, as in reality there are no known tissues where this is the case. 
The details of the simulation can be found in \cite{Cruz2018}.

We compared our algorithm using LSTM and GRU with the following techniques on the magnitude of the complex MRF signals: 1) ANN: Two fully connected layers of 300 neurons as described in \cite{Cohen2018}; 2) 1D CNN: Convolutions applied on the 1D signal as described in \cite{Hoppe2017}; 3) Inner product: exhaustive inner product multiplication with each signal in the dictionary. 

We trained all methods on the magnitude values of 100000 signals and tested on 5000 signals. In Table \ref{tab:Quan}, we report the mean absolute error (MAE) and root mean square error (RMSE) between our predicted T1/T2 maps and the gold standard maps. The results show that even a simple RNN-based architecture is capable of predicting the T1 and T2 values with high accuracy compared to other techniques. In particular, deep learning based techniques can handle noise better and generate results much faster. The inherent temporal information in the MRF signal has generated high accuracy for both LSTM and GRU architectures. GRU gives better results compared to LSTM, both for T1 and T2. This can explained by the smaller number of parameters that are needed to be learned by GRUs. In the literature, \cite{Yin2017} also reported superior performance of GRUs for a speech signal modeling task, similar to our problem.

\section{Discussion and Conclusions}
\label{sec:Discussion}

In this paper we have proposed a RNN-based technique for reconstructing T1 and T2 maps, and demonstrated it using a synthetic MRF dataset.  
We have shown that even a simple RNN architecture is capable of predicting T1 and T2 values with considerable success, outperforming other state of the art techniques. We believe that this is due to their ability to learn data-driven time-dependent features. MR parameter variation in MRF signal generation is continuous and the pattern repeats through time. 
GRU and LSTM are capable of capturing this redundant information. 

MRF can be extended to other parameters, such as B0, B1, MT (magnetization transfer), diffusion, etc. and the scalability of the reconstruction technique is a bottleneck for these applications.  The dictionary will grow exponentially and explode quickly for these truly multi-parametric applications. We have shown that RNN architectures are capable of producing accurate reconstructions (of T1 and T2) in less than 20ms once they are trained for an hour. Similar results may be achievable for these other parameters, and we will investigate this in future work.

One important aspect is the undersmpled highly-undersampled k-space in MRF, which should be also accounted for in future applications of neural networks. We would also like to validate our method on phantom and in-vivo MRF acquisitions, and include spatio-temporal features into our model based on spatial correspondences from in-vivo MRI acquisitions.

\bibliographystyle{IEEEbib}
\bibliography{ISBI_MRF}

\begin{thebibliography}{10}

\bibitem{Ma2013}
Dan Ma, Vikas Gulani, Nicole Seiberlich, Kecheng Liu, Jeffrey~L Sunshine,
  Jeffrey~L Duerk, and Mark~A Griswold,
\newblock ``Magnetic resonance fingerprinting,''
\newblock {\em Nature}, vol. 495, no. 7440, pp. 187, 2013.

\bibitem{Coppo2016}
Simone Coppo, Bhairav~B Mehta, Debra McGivney, Dan Ma, Yong Chen, Yun Jiang,
  Jesse Hamilton, Shivani Pahwa, Chaitra Badve, Nicole Seiberlich, et~al.,
\newblock ``Overview of magnetic resonance fingerprinting,''
\newblock in {\em MAGNETOM Flash}, p.~65. 2016.

\bibitem{Wang2014}
Z.~Wang, Q.~Zhang, J.~Yuan, and X.~Wang,
\newblock ``{MRF} denoising with compressed sensing and adaptive filtering,''
\newblock in {\em {IEEE} 11th International Symposium on Biomedical Imaging
  ({ISBI})}, 2014, pp. 870--873.

\bibitem{Gomez2015}
Pedro~A. G{\'{o}}mez, Cagdas Ulas, Jonathan~I. Sperl, Tim Sprenger, Miguel
  Molina{-}Romero, Marion~I. Menzel, and Bjoern~H. Menze,
\newblock ``Learning a spatiotemporal dictionary for magnetic resonance
  fingerprinting with compressed sensing,''
\newblock in {\em {MICCAI} Workshop on Patch-Based Techniques in Medical
  Imaging ({Patch-MI})}, 2015.

\bibitem{Cohen2018}
Ouri Cohen, Bo~Zhu, and Matthew~S Rosen,
\newblock ``{MR} fingerprinting deep reconstruction network ({DRONE}),''
\newblock {\em Magnetic resonance in medicine}, vol. 80, no. 3, pp. 885--894,
  2018.

\bibitem{Hoppe2017}
Elisabeth Hoppe, Gregor K{\"{o}}rzd{\"{o}}rfer, Tobias W{\"{u}}rfl, Jens Wetzl,
  Felix Lugauer, Josef Pfeuffer, and Andreas~K. Maier,
\newblock ``Deep learning for magnetic resonance fingerprinting: {A} new
  approach for predicting quantitative parameter values from time series,''
\newblock in {\em German Medical Data Sciences: Visions and Bridges -
  Proceedings of the 62nd Annual Meeting of the German Association of Medical
  Informatics, Biometry and Epidemiology}, 2017.

\bibitem{Virtue2017}
Patrick Virtue, Stella~X. Yu, and Michael Lustig,
\newblock ``Better than real: Complex-valued neural nets for {MRI}
  fingerprinting,''
\newblock in {\em {IEEE} International Conference on Image Processing
  ({ICIP})}, 2017, pp. 3953--3957.

\bibitem{Balsiger2018}
Fabian Balsiger, Amaresha~Shridhar Konar, Shivaprasad Chikop, Vimal Chandran,
  Olivier Scheidegger, Sairam Geethanath, and Mauricio Reyes,
\newblock ``Magnetic resonance fingerprinting reconstruction via spatiotemporal
  convolutional neural networks,''
\newblock in {\em {MICCAI} Workshop on Machine Learning for Medical Image
  Reconstruction ({MLMIR})}, 2018.

\bibitem{Fang2018}
Zhenghan Fang, Yong Chen, Mingxia Liu, Yiqiang Zhan, Weili Lin, and Dinggang
  Shen,
\newblock ``Deep learning for fast and spatially-constrained tissue
  quantification from highly-undersampled data in magnetic resonance
  fingerprinting {(MRF)},''
\newblock in {\em {MICCAI} Workshop on Machine Learning in Medical Imaging
  ({MLMI})}, 2018.

\bibitem{Hochreiter1997}
Sepp Hochreiter and J{\"u}rgen Schmidhuber,
\newblock ``Long short-term memory,''
\newblock {\em Neural computation}, vol. 9, no. 8, pp. 1735--1780, 1997.

\bibitem{Cho2014}
Kyunghyun Cho, Bart Van~Merri{\"e}nboer, Dzmitry Bahdanau, and Yoshua Bengio,
\newblock ``On the properties of neural machine translation: Encoder-decoder
  approaches,''
\newblock {\em arXiv preprint arXiv:1409.1259}, 2014.

\bibitem{Hennig2004}
Juergen Hennig, Matthias Weigel, and Klaus Scheffler,
\newblock ``Calculation of flip angles for echo trains with predefined
  amplitudes with the extended phase graph (epg)-algorithm: principles and
  applications to hyperecho and traps sequences,''
\newblock {\em Magnetic Resonance in Medicine: An Official Journal of the
  International Society for Magnetic Resonance in Medicine}, vol. 51, no. 1,
  pp. 68--80, 2004.

\bibitem{Cruz2018}
Gast{\~a}o Cruz, Olivier Jaubert, Torben Schneider, Rene~M Botnar, and Claudia
  Prieto,
\newblock ``Rigid motion-corrected magnetic resonance fingerprinting,''
\newblock {\em Magnetic resonance in medicine}, 2018.

\bibitem{Yin2017}
Wenpeng Yin, Katharina Kann, Mo~Yu, and Hinrich Sch{\"u}tze,
\newblock ``Comparative study of cnn and rnn for natural language processing,''
\newblock {\em arXiv preprint arXiv:1702.01923}, 2017.

\end{thebibliography}

\end{document}